\documentclass[letterpaper, 10 pt, conference]{ieeeconf}  

\IEEEoverridecommandlockouts                              

\usepackage{graphicx} 
\usepackage{amsmath} 
\usepackage{amssymb}  

\usepackage{color}
\usepackage{algorithm2e}
\usepackage{graphicx} 
\usepackage{float}
\usepackage{fancyvrb}
\usepackage{pifont} 
\newcommand{\meta}[1]{}

\title{\LARGE \bf
A Model-Driven Engineering Approach for ROS\\using Ontological Semantics
}

\author{Stefan Zander$^{1}$ and Georg Heppner$^{1}$ and Georg Neugschwandtner$^{2}$ and Ramez Awad$^{3}$ and \\ Marc Essinger$^{1}$ and Nadia Ahmed$^{1}$ 
\thanks{$^{1}$Stefan Zander, Georg Heppner, Marc Essinger, and Nadia Ahmed are with the FZI Research Center for Information Technology, Haid-und-Neu-Str. 10--14, 76131~Karlsruhe, Germany
        {\tt\small \{lastname\}@fzi.de}}%
\thanks{$^{2}$Georg Neugschwandtner is with fortiss GmbH, Guerickestr. 25, 80805 Munich, Germany
        {\tt\small neugschwandtner@fortiss.org}}%
\thanks{$^{3}$Ramez Awad is with the Fraunhofer Institute for Manufacturing Engineering and Automation (IPA), Nobelstr. 12, 70569 Stuttgart, Germany
        {\tt\small ramez.awad@ipa.fraunhofer.de}}%
}

\begin{document}

\maketitle
\thispagestyle{empty}
\pagestyle{empty}

\begin{abstract}
This paper presents a novel ontology-driven software engineering approach for the development of industrial robotics control software. 
It introduces the ReApp architecture that synthesizes model-driven engineering with semantic technologies to facilitate the development and reuse of ROS-based components and applications.
In ReApp, we show how different ontological classification systems for hardware, software, and capabilities help developers in discovering suitable software components for their tasks and in applying them correctly.
The proposed model-driven tooling enables developers to work at higher abstraction levels and fosters automatic code generation. It is underpinned by ontologies to minimize discontinuities in the development workflow, with an integrated development environment presenting a seamless interface to the user.
First results show the viability and synergy of the selected approach when searching for or developing software with reuse in mind.
\end{abstract}
\section{INTRODUCTION}

While traditionally industrial robots were only used in mass-production facilities to automate specialized but highly repetitive tasks, more and more flexibility is being called for across all industrial sectors to meet the growing demand for fast, adaptable, and accurate production processes~(cf.~\cite{gruninger2009market}): Small and medium-sized enterprises (SMEs) are increasingly seeking to rely on robots within their production processes. 
The transition from repeating the same process thousands or even millions of times over towards flexible, but still efficient production of small lot sizes and varying products, all the way down to full customization at lot size one, requires new efficient software development approaches. 

Today, a wide range of proprietary robot control languages and system environments exist that restrict the reuse of developed software---often to a single application scenario. They require experts with detailed system knowledge to reconfigure and extend the production software.
Therefore, increasing their reusability is an essential step towards creating robotics software applications more efficiently. Reusability can be increased in a number of ways, in particular:
\begin{itemize}
\item Writing software as loosely-coupled components~\cite{szyperski:2002}
\item Fostering standardization of execution environments for these components 
\item Establishing standard interfaces (both in terms of platform/protocol and data types) for standard tasks
\item Facilitating component retrieval and exchange 
\end{itemize}

ROS~\cite{quigley2009ros} already goes a long way towards meeting these requirements. It provides a software framework that allows the reuse of software components in various applications and provides a convenient infrastructure for generic tasks like communication, component discovery and even the user interface.
The highly active ROS community provides a large number of readily available software packages to tackle many recurring problems in robotics such as image recognition, control or motion planning. However, the wealth of available packages (currently over 1900) leads to new problems: How can a user decide which package is the right one for their application? And how does one contribute software in a way that the reusability in other projects can be assured?

The main objective of ReApp is to further improve the re-use of robotics software. To this end, ReApp builds upon the ROS component model, but introduces significant enhancements in terms of metadata and tooling. A model-driven design methodology is backed by a semantic description of software components based on ontologies. The ReApp Workbench (cf.\ Section~\ref{sub:tool_chain} and Figure~\ref{fig:reapp_overall}), an integrated development environment for implementing, assembling, annotating, and deploying software components, works in conjunction with a cloud-based semantic repository, the ReApp Store~\cite{bastinos:2014}, to support the needs of different user roles (component developers, system integrators, etc.) during all stages of software development.
The underlying hypothesis is that \emph{such a combination of methodologies, technologies, and tools significantly improves means that foster discovery and reusability of ROS components}. 

To address the problem of finding suitable software for a particular task, ReApp software components, referred to as \emph{Apps}, do not only carry the interface information that describes the technicalities of data exchange with other components (such as available ROS topics or parameters), but also high-level semantic information about their use in a robotics application, based on the ReApp ontologies (see Section~\ref{sub:ontologies}). This allows, for example, a system integrator to search for a software component by merely providing a broad categorization of what a component should be used for, by specifying certain capabilities the component provides, by selecting a particular hardware device that a driver component should support, or even by asking for a component that will fit with other selected components. We present a concrete example in Section~\ref{sub:using_software}.

To be able to provide this kind of assistance, the ReApp Workbench and Store require basic information about the Apps, which has to be collected from the component developers. The ReApp Workbench is designed to make it convenient for a software or hardware developer to specify relevant attributes such as desired capabilities, or simply a general concept of the software they are developing. \meta{TODO: We can either add a screenshot of the CMT Tool here or add it to Sec III and place a forward reference here.}This process is guided by the ReApp ontologies which are aligned with the model-driven part of the ReApp Workbench. This allows the Workbench to create high-level models of components and their interfaces---which are later transformed to source code---automatically and to collect all relevant information from the developer in a single, integrated step.

\section{RELATED WORK}

\subsection{Software Architecture and Robotic Components}
The complexity of robotic software systems can only be mastered with well-designed software architectures and integrated IT tool chains that support the entire development process (see~\cite{kortenkamp2008robotic}). 
The most widely used software frameworks within robotics (cf.~\cite{elkady2012robotics}) are the Robot Operating System (ROS)\footnote{www.ros.org}~\cite{quigley2009ros}, Orocos\footnote{www.orocos.org}~\cite{ioris2012integrating}, and OpenRTM\footnote{www.openrtm.org}~\cite{ando2008software}. 
The ROCON\footnote{Robotics In Concert: http://www.robotconcert.org/} project aims to simplify the configuration, retaskability, and reuse of robotic systems by iterating on ROS. Each robot can execute different task-driven robot apps that define all required components to fulfill a specific task. 
Currently, these frameworks are primarily used in the service robotics domain, so that only a few industrial sample applications exist. This is, among others, due to the lack of quality standards for those frameworks, especially with respect to deterministic behavior and real-time functionalities~\cite{angerer2010robotics}. However, in addition to these quasi-standards, a variety of more locally-distributed robotics frameworks such as SmartSoft or MCA2~\cite{uhl2007mca2} exist. Only with the establishment of the ROS Industrial Initiative, a concentrated effort to develop an industrial-grade software framework was undertaken~\cite{edwards2012ros}.

\subsection{Model-driven Engineering of Robotic Systems}
\ifx true false
\meta{IEC 61499 is about modelling, components (and more), but I would only refer to a comparatively minor part of the IEC 61499 \emph{tool chains} as model-driven. Maybe the subsection heading can be improved (GN).}
\meta{Even so, i see no problem in mentioning it as it at least follows some principles of the approach. With respect to time and effort i would just leave it exactly as it is (GH).}
\meta{Alternative title could be? Model-based approaches to Engineering of Robotic Systems ME}
\fi

Schlegel et al.~\cite{schlegel2006} recognize the need for adopting engineering principles in order to cope with complexity in robotic software systems. Their SMARTSOFT meta model allows robotic software components to be modeled using UML, clearly defining their external interface using a constricted set of interaction patterns and modelling their execution using a state automaton. The resulting platform independent models are then transformed into code templates. 

Within the BRICS project~\cite{bruyninckx2013brics}, a component model was defined ``to provide developers with as much structure in their development process as is possible without going into any application-specific details''. It follows the model abstraction levels defined by Object Management Group (OMG)\footnote{http://www.omg.org/} 
and uses a Component-Port-Connector (CPC) model to represent the software component at M2 level. To aid developers in the transition from the platform-independent model to the platform-specific model and ultimately to the actual implementation of the software component, BRIDE (an Eclipse Plug-In) was developed, providing model-to-model transformation and code generation capabilities~\cite{bubeck2014bride}.

IEC 61499 is an open standard for modelling distributed control and automation systems. An IEC 61499 system consists of one or more control devices, each hosting one or more software execution environments, so-called resources. IEC 61499 applications are built from function blocks connected by event and data flows and can span multiple resources. Function blocks are typed, reusable software components with a well defined interface. An IEC 61499 IDE such as 4DIAC\footnote{http://www.eclipse.org/4diac}~\cite{zoitl2013reusablecontrol} will transform a formal function block type definition into a source code skeleton. 

\subsection{Using Ontologies in Robotics}

A detailed discussion about the beneficial interaction between ontologies and model-driven approaches was published in~\cite{assmann2006ontologies}.
In this context,~\cite{martin2007bringing} demonstrates the semantic modelling of Web service composition and their functionalities through task-specific ontologies and predefined components. This approach is similar to the modelling of robotic components and their interaction in our work.
The use of ontologies in guiding component composition has also been demonstrated with the LARKS system~\cite{sycara2002larks} that matches web-based software agents to each other. In~\cite{gil2001phosphorus} agents and tasks are modeled and matched to each other by subsumption, a specific type of expressing generality and specialization entailments. Compton et al.~\cite{compton2009reasoning} have developed an ontology using the Web Ontology Language (OWL) for robots, the task of which is search and rescue in urban areas. A similar approach is described in~\cite{nilsson2009ontology}; there, the assignment of tasks to specific robot components is left to the user rather than computed by formal reasoning on logical entailments. The representation of sensorial information and observables is another scenario for using ontologies in robotics. Standardization activities towards unified sensor specifications are driven by the W3C Semantic Sensor Network Incubator Group\footnote{http://www.w3.org/2005/Incubator/ssn/}. The state-of-the-art regarding the semantic specification of sensors is summarized in~\cite{compton:2009a}. 

ReApp aims at enhancing selected aspects of the aforementioned works and integrates them into a single tool chain (see Section~\ref{sub:tool_chain}) that supports different development roles. In addition to semantic descriptions of interfaces, ReApp components also model expert knowledge regarding their usage and deployment. ReApp employs a capability-based concept for computing compatibilities among components through the notion of reasoning on a model's formal semantics and facilities its discovery using semantic search tools. ReApp uses the Semantic Sensor Network Ontology\footnote{http://www.w3.org/2005/Incubator/ssn/ssnx/ssn} and employs mechanisms for the deduction of capabilities from interface specifications.
\section{APPROACH}

ReApp follows a model-driven approach to the development of robotic components/applications. 
It specifies an underlying information model (see~\ref{sub:information_model}), the manifestation of which is represented using formal description frameworks defined as part of the W3C's Data Activity standardization efforts\footnote{See http://www.w3.org/2013/data/} in combination with domain-specific semantics defined in three different main ontologies for hardware, software and capability-related information (see~\ref{sub:ontologies}).
This enables a unified, detailed, and coherent description of all relevant component elements.

In ReApp, robotic components and applications are structured as \emph{Apps}. An App is the unit of software distribution in ReApp: An App is a reasonably self-contained piece of software with a well-defined interface and purpose that can be downloaded from the \emph{ReApp Store}. An App can either be a 
\emph{Hardware Access Component \mbox{(HA-}Component)} that interfaces with a device, a \emph{``pure''\footnote{``Pure'' in the sense that a HA-Component is a software component as well, but one that is associated with a particular hardware device.} Software Component \mbox{(SW-}Component)} that provides computation or control functions which are not specific to a particular hardware device (such as detecting objects in images), or a combination of these components.
Such a combination or assembly of components usually defines higher-level application functionality. Therefore, it is referred to as a \emph{Skill} in ReApp.

Skills ``wire'' Apps together, defining the binding between the Apps' communication interfaces. Since Skills are themselves Apps, a hierarchy of component assemblies can be defined by nesting Skills as appropriate. To handle process-related sequence logic, Skills can contain SW-Components dedicated to this purpose, which are called \emph{Coordinators}.
A Skill can be parameterized to form a \emph{Solution}, a term used in ReApp to describe a component assembly that is ready for deployment and is provided with all necessary information (such as configuration files or calibration data) to start executing on the target hardware.

\begin{figure}[ht]
	\centering
	\includegraphics[width=\linewidth]{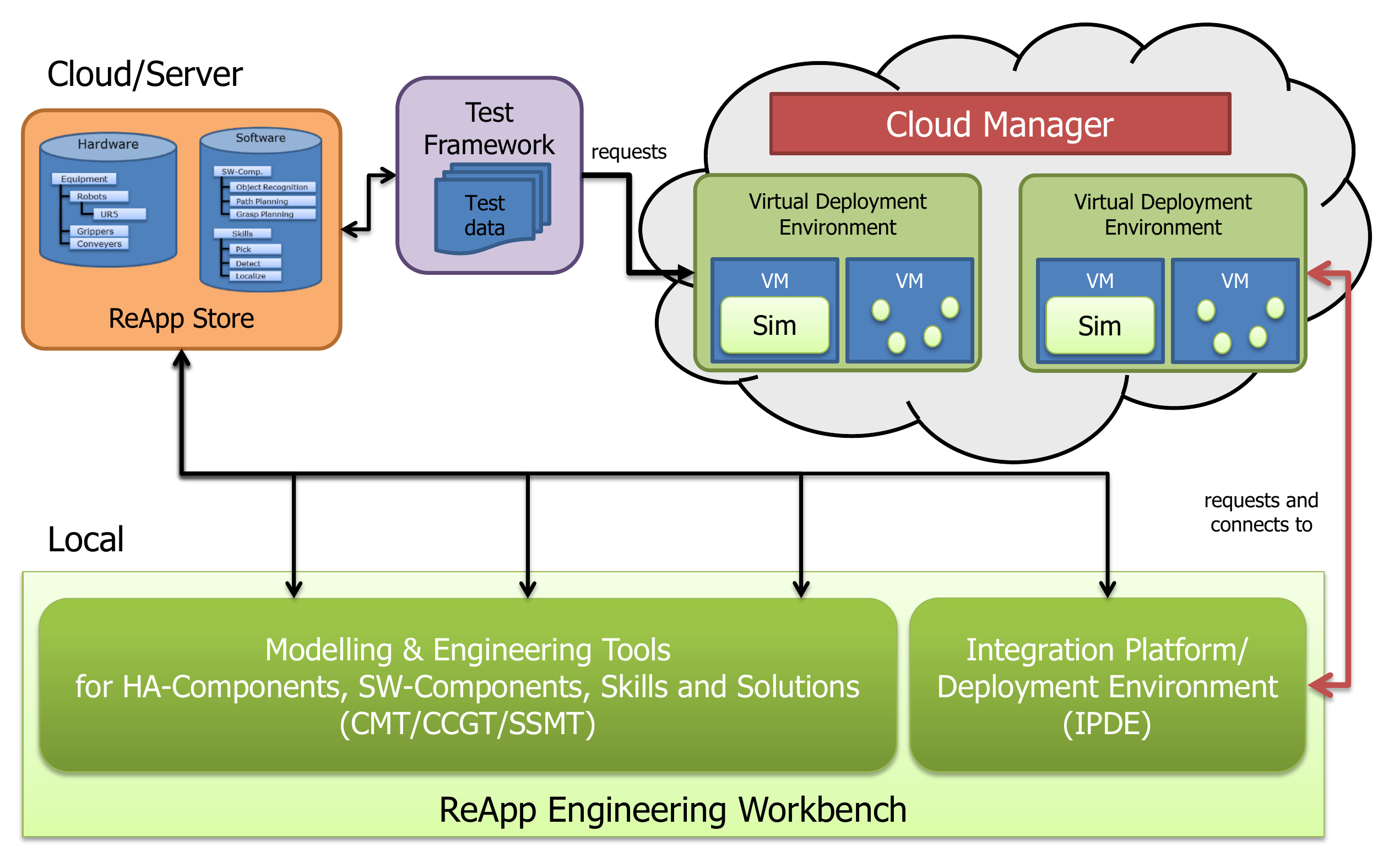} 
	\caption{The tool and systems infrastructure of ReApp including the ReApp Engineering Workbench, the ReApp Store and cloud services} 
	\label{fig:reapp_overall}
\end{figure}

Figure~\ref{fig:reapp_overall} provides an overview of the ReApp system concept. ReApp targets the entire lifecycle of software components, reaching from modelling over coding towards assembly and execution. Cloud-based execution of components is an integral part, at production time on one hand (if timing requirements are sufficiently relaxed) as well as at development time for automated testing. Section~\ref{sub:tool_chain} briefly describes the model-driven engineering tools for \mbox{HA-}Components, \mbox{SW-}Components and Skills, whose semantics are the focus of this paper.

\subsection{Tool Chain} 
\label{sub:tool_chain}

\begin{figure}[ht]
	\centering
	\includegraphics[width=0.9\linewidth]{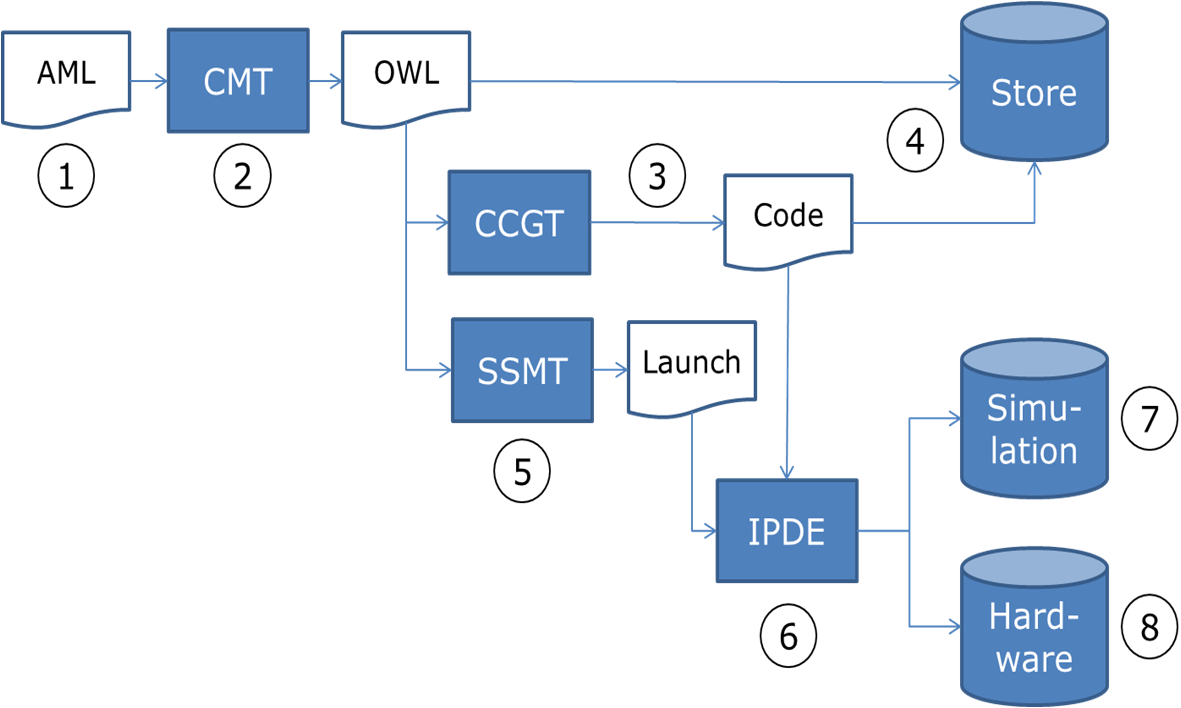} 
	\caption{Data flow through the ReApp tool chain} 
	\label{fig:reapp_workflow}
\end{figure}

Figure~\ref{fig:reapp_workflow} outlines how the parts of the ReApp tool and systems infrastructure shown in Figure~\ref{fig:reapp_overall} work together to support creating and deploying an application and all relevant component models. \meta{TODO: Please check and correct if necessary}
Rectangles (with dark background) refer to constituting tools of the ReApp Engineering Workbench. Cylinders stand for other parts of the ReApp infrastructure, which are not within the scope of this paper.
Document shapes (with light background) represent artefacts such as created models or files that are exchanged between the involved tools; consequently, arrows indicate data flows and logical sequence.

\begin{figure}[ht]
	\centering
	\includegraphics[width=0.9\linewidth]{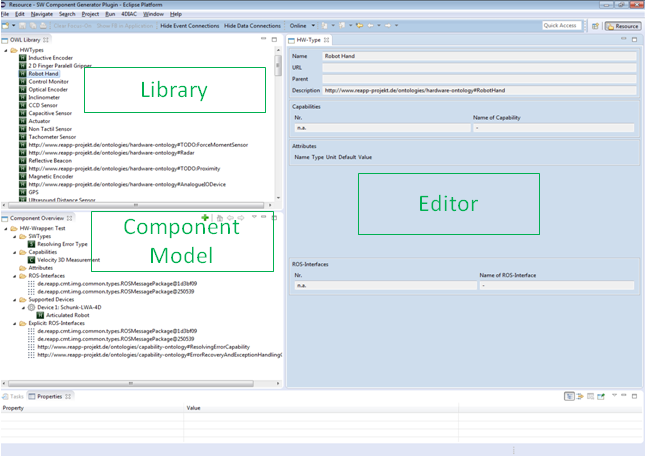} 
	\caption{The Component Modelling Tool}
	\label{fig:reapp_cmt}
\end{figure}

Component developers first build a local model of their intended robotic component, specifying its capabilities and interfaces. 
They are aided by the \emph{Component Modelling Tool (CMT \ding{193}, Fig.~\ref{fig:reapp_cmt})}, which provides model element libraries as well as an import and conversion facility (\ding{192}) for component interface descriptions represented in the Automation Markup Language~\cite{drath:2008a,hundt:2008}.
This option was included in order to ease the creation of HA-Components by parsing AutomationML descriptions provided by manufacturers.
The resulting model is stored as an RDF/OWL-file,\footnote{See http://www.w3.org/TR/owl2-xml-serialization/} which can then be used for further processing steps or stored in a component repository, the ReApp Store (\ding{195}).

Next, the \emph{Component Code Generation Tool (CCGT \ding{194})} parses the model and generates an executable code skeleton with all the specified interfaces and ROS-specific information (i.e.~the \textsf{package.xml}).
The language of the generated code depends on user input as well as the target platform. 
Developers are then required to implement the actual functionality using the selected language. Code generation is currently ongoing work; support for C++ will be provided, and the integration of IEC 61499 function block networks is planned.
Once the developer has finished the implementation, the component's executables and its model can be uploaded to the ReApp Store (\ding{195}) in form of a ready to use App.

\begin{figure}[ht]
	\centering
	\includegraphics[width=0.9\linewidth]{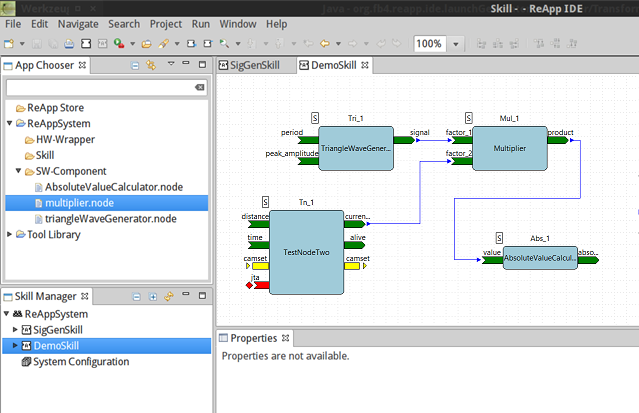} 
	\caption{The Skill/Solution Modelling Tool}
	\label{fig:reapp_ssmt}
\end{figure}

System integrators make use of these Apps by searching the ReApp Store to find the functionality they require in a project. They then download the models of those Apps and assemble them into Skills by connecting their interfaces. They may also add locally created Components to a Skill, in particular Coordinators to sequence the triggering events of the Components within the Skill. They will also want to set parameters for the Skill's Components, and, if the Skill is to be uploaded to the Store, review and edit the relevant Skill model information. All these activities are supported by the \emph{Skill/Solution Modelling Tool (SSMT~\ding{196}, Fig.~\ref{fig:reapp_ssmt})}, which provides a graphical (``block diagram'') editor. When assembling a Skill, the SSMT can also provide assistance based on the Apps' semantic annotations.  

\meta{The following was/is pretty obscure -- please check and correct if necessary. What is the ``matching information in the Apps' model''? -gn} To actually complete, test and execute a Solution, ReApp provides the \emph{Integration Platform \& Deployment Environment} (\ding{197}), which has two main functionalities: 
\begin{enumerate}
	\item Associating all HA-Components with either their corresponding hardware devices (\ding{199}) or alternatively their representations in the simulation environment (\ding{198})---using the matching information in the Apps' model to parameterize the simulation model of the hardware device---and assigning all SW-Components to a suitable execution target environment (local or in the cloud).
	\item Deploying all Apps, i.e.\ managing their download to the proper execution target, establishing the correct communication ``plumbing'' between them and triggering their execution (by way of the top-level Coordinator, if appropriate).
\end{enumerate}

\subsection{Information Model} 
\label{sub:information_model}

As was discussed before, the parts of the ReApp Workbench rely on the information model for Apps for exchanging App-related information. More importantly, the model is used in conjunction with the ontologies to enable model-based search and matching, improve interchangeability among Apps, and make automated functional quality tests and reports possible. 

The information model for Apps in ReApp consists of the following baseline elements:
\begin{figure*} 
	\centering
	\includegraphics[width=\textwidth]{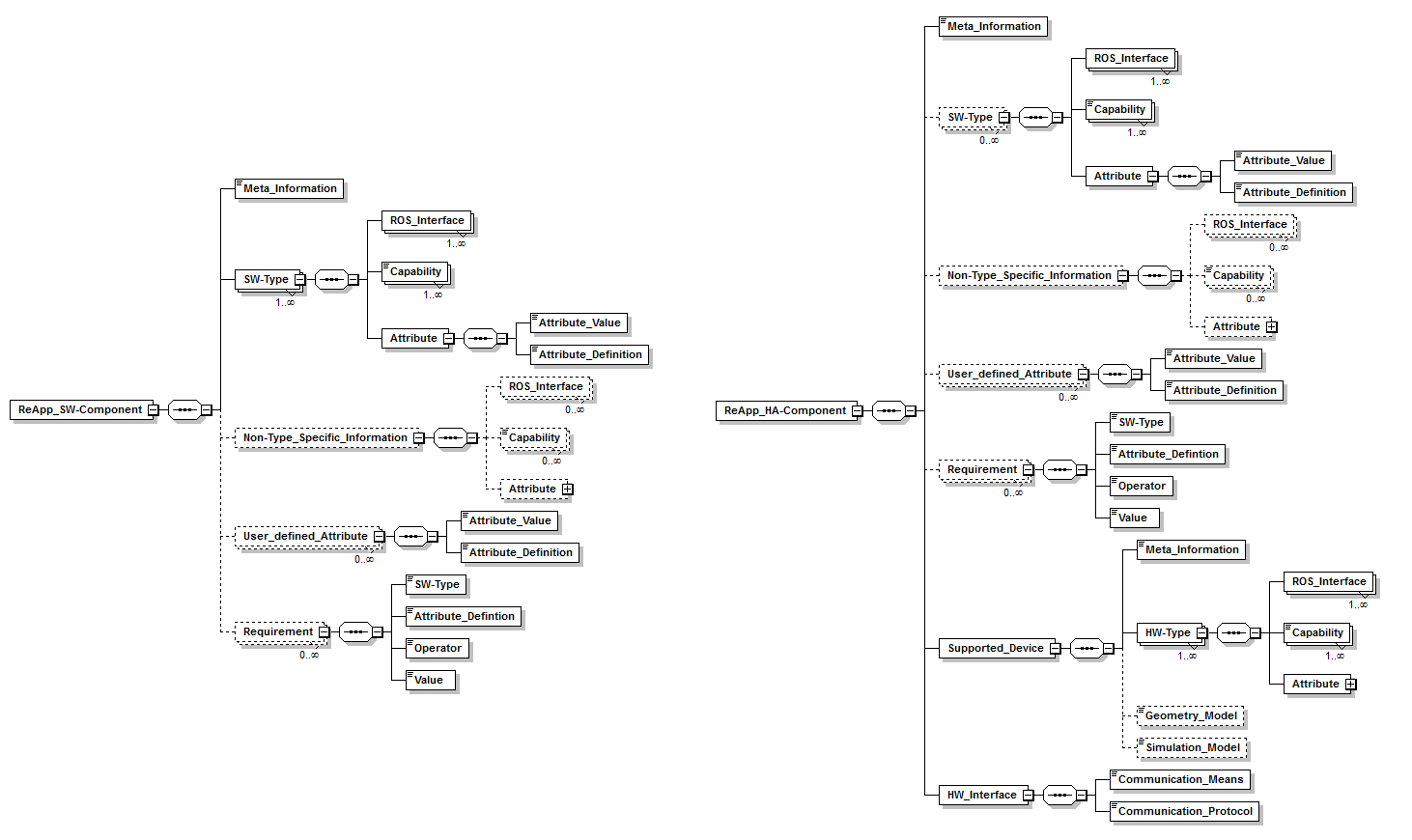} 
	\caption{ReApp Model of ``pure'' Software Component (left) and Hardware Access Component (right)}
	\label{fig:component_models}
\end{figure*}
\begin{itemize}
	\item \textbf{Meta Information:} This includes information such as the author of the App, its owner, its creation time, its current version, whether the app is a SW-Component or a HA-Component, etc. 
	It also contains a textual description of the purpose and the functionality of an App, intended primarily for users browsing through the ReApp Store.
	
	Additionally, it contains the status of the App, indicating whether it is just a model, a prototype of the App (with executables that have not undergone automated testing yet) or a fully tested and released App that can be reliably downloaded by others and used in their Skills/Solutions. 
	
\item \textbf{SW-Types:} A SW-Type is a specified bundle of capabilities, ROS interfaces (i.e. provided or requested communication endpoints: Topics, Actions, Services \footnote{See http://wiki.ros.org/ROS/Patterns/Communication}) and attributes associated with a certain type of software components\footnote{ It should be noted that a pure software component has at least one SW-Type. 
It may have several SW-Types. A HA-Component may have a SW-Type but it is not required to have one.}.
ReApp specifies a taxonomy of SW-Types, which the developer can assign to their App's model.
By assigning a certain SW-Type to the App, the developers are stating that the App will support the ROS interfaces associated with that SW-Type; that it will provide the capabilities expected by such a SW-Type; and that the App will also have specific attributes listed in the SW-Type, which may be used to filter the search results in the ReApp Store.
Example: The App \textsf{RaVision} has the SW-Type \textsf{Detect\_Object\_in\_Image}.
Associated with that SW-Type is the capability of detecting whether an object is in a specific image or not.
It is expected to subscribe to a ROS Topic \textsf{sensor\_msgs/Image} and to publish to a ROS Topic \textsf{object\_detection/Object\_detected}.
\item \textbf{Non-Type Specific Information:} 

To differentiate themselves from existing Apps, Apps occasionally employ additional capabilities, interfaces, and attributes that go beyond the default specifications for their assigned SW-Types, but fall short of fulfilling the requirements of an additional SW-Type.	
For this reason, the ReApp model can include Non-Type-Specific Information, where the developer can add additional capabilities, ROS interfaces and attributes from the ReApp taxonomies,~and which is also taken into account for semantic indexing and retrieval. 
\item \textbf{Requirements:} The developers can define requirements on other Apps to be connected with their App in Skills/Solutions. 
For example, the App \textsf{RaVision} of the type \textsf{Detect\_Object\_in\_Image} may require that it is connected with another App of the type \textsf{RGBD-Camera\_Wrapper} that provides at least 30 frames per second (FPS)---otherwise \textsf{RaVision} cannot guarantee its capability to reliably detect objects. 
A requirement is modelled by specifying the type of the other App and the relevant type's attribute; then the constraint to the value of that attribute is given as a formula, e.g.\ \textsf{RGBD-Camera\_Wrapper.FPS {\textgreater} 30.0}. This is done using a wizard in the CMT (support for more complex logic expressions is ongoing work).
These requirements can be processed by the SSMT to prevent it from suggesting Apps that are not suitable for being connected to \textsf{RaVision} and/or to display a warning when the system integrator tries to manually connect an unsuitable App.
\end{itemize}
The model of a HA-Component has additional elements that are important for controlling the hardware device(s). These are:

\begin{itemize}
\item \textbf{Supported Devices:} This specifies the actual hardware devices that this App has been implemented, tested and released for\footnote{A HA-Component can be implemented in a generic way to control different hardware devices of the same type; e.g., a ROS-driver can be used for several different robot arms.}. 
It contains meta-information about the device, like manufacturer, model name and number. 
It also contains the HW-Type of those supported devices. Like SW-Types, HW-Types specify a bundle of capabilities, ROS-interfaces, and attributes; however, they do so with respect to a certain hardware device type. 
Additionally, the developer may choose to include a geometry model of the device and/or its simulation model. 
This information may later be used in the SSMT or the IPDE for testing the application.
\item \textbf{HW-Interfaces}: Situated at a lower level of abstraction than the ROS interfaces, the HW-interfaces specify the communication means (USB/RS232/Ethernet etc.) and protocol (EtherNet/IP, CANopen, DeviceNet etc.) to be used for interfacing with the actual hardware device and for triggering its functions. 
This information can be used in the CCGT to generate a code skeleton for that interface as well. 
\end{itemize} 

\ifx true false
\meta{RFA: bei bedarf kann man den restlichen Absatz kuerzen}
For example, using the information about the app \textsf{RaVision} a test node can be started which sends a series of images to \textsf{RaVision}. 
This node subscribes to the topic \textsf{object\_detection/Object\_detected} and compares the published results with its stored ground truth data. 
At the end of the test it is able to compile a report stating the percentage of positive object detections. 
This can be done using different image series, varying the object and illumination properties, thus compiling a more thorough report, about the performance of \textsf\textsf{RaVision} under different parameter settings. 
This report may be consulted by a system integrator when choosing which app of the type \textsf{Detect\_Object\_in\_Image} to download and purchase.
\fi

\subsection{Ontologies} 
\label{sub:ontologies}
\begin{figure*}
	\centering
	\includegraphics[width=0.8\textwidth]{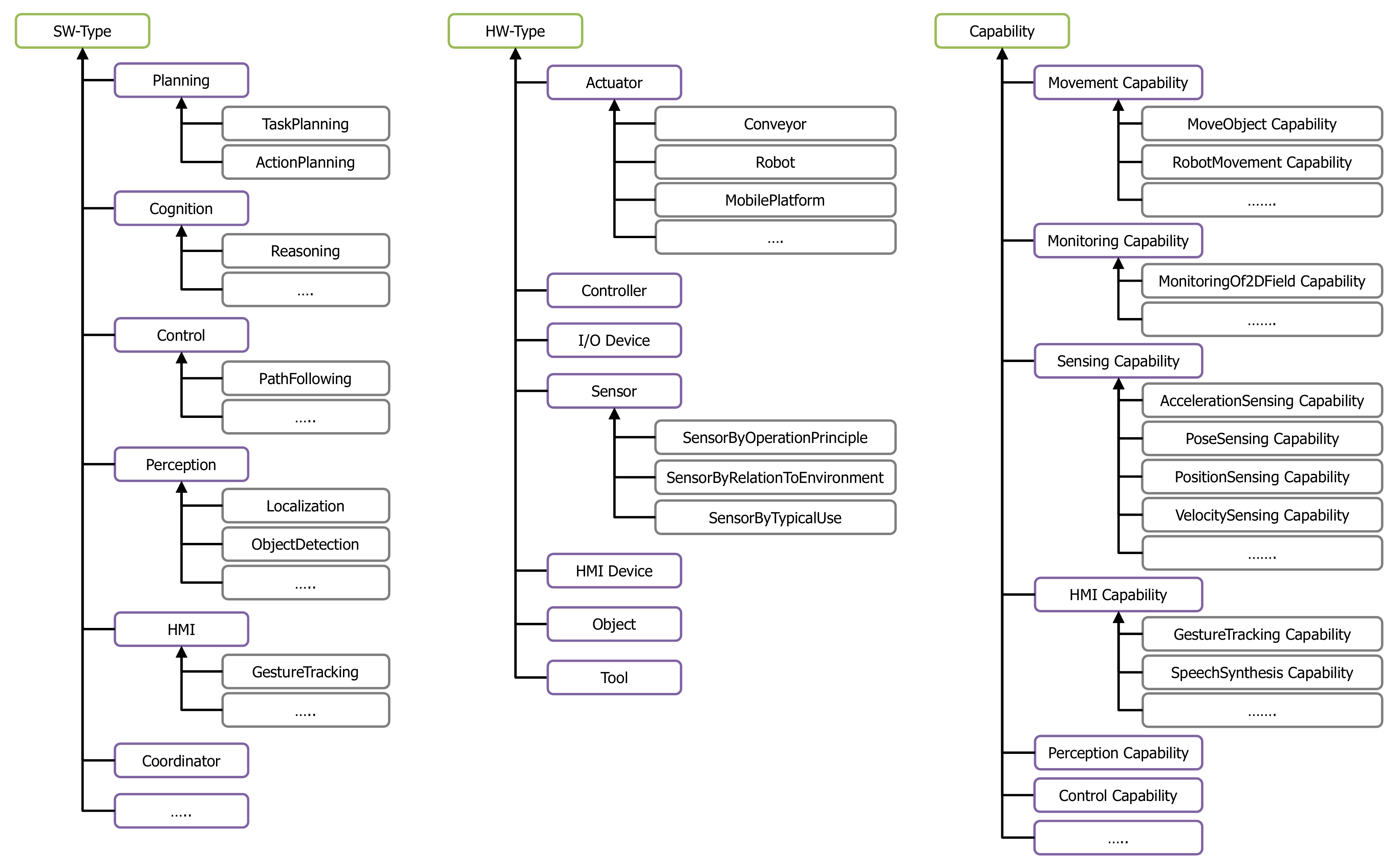} 
	\caption{Excerpt of the top-level classification systems implemented in the ReApp Hardware, Software and Capability Ontologies}
	\label{fig:ontology}
\end{figure*}

In the model-driven approach proposed by ReApp, ontologies serve as the glue that integrates the different model and information types (see Section~\ref{sub:information_model}). The main reason why we propagate the use of ontologies as data interoperability infrastructure is that they ensure \emph{data uniformity}, i.e., they allow the description of data semantics in an application-independent way. Ontologies in general provide a number of advantages wherefore they were our first choice in representing ROS-related information:
(i) ontologies allow the formal and explicit representation of data semantics upon which the deduction of logical entailments from given axiom sets can be computed by a reasoner. This is a main feature that distinguishes ontologies from most commonly used modeling languages in Information Science such as the UML~\cite{kroetzsch:2014a}.
(ii) Semantic Web ontologies are formally built on decidable fragments of first-order logic, the so-called family of \emph{Description Logics} that allow humans and machines to formally describe and exchange information with little to no ambiguity (see~\cite{kroetzsch:2014a,baader:2003,gil:2005,rudolph:2011,zander:2015}).
(iii) A reasoning engine can be used to process the formal model-theoretic semantics of HA- and SW-Components and deduce implicit knowledge through the notion of logical consequence\footnote{See~\cite{zander:2015} for information about the type of reasoning problems relevant for the ReApp project that can be answered by a default DL reasoner.}. This allows us the implementation of some model-checking algorithms in which the reasoner checks the satisfiability of constituting modeling axioms of components or a configuration and thus answers whether a component description is consistent or not.
The following sections introduce the different types of ontologies used in ReApp and illustrate their top level-elements (see Figure~\ref{fig:ontology}). 

\subsubsection{Hardware Ontology}
\label{ssub:hardware_ontology}
The hardware ontology defines a classification system for hardware devices such as sensors, actuators, controllers etc.\ and allows for a categorization of such devices along different dimensions, the memberships of which are computed by a reasoner (see~\cite{zander:2015} for details and examples). It also defines concrete terms for the description of device features and characteristic elements and contains a taxonomical classification system for communication interfaces and the protocols being used. In general, the hardware ontology exhibits different classification levels:
\begin{itemize}
	\item Level 0: Represents the root concept of the hardware ontology, i.e., \textsf{Hardware Type}
	\item Level 1: Defines general hardware types for the components of a robot system such as \textsf{Sensor}, \textsf{Controller}, \textsf{Actuator} etc.
	\item Level 2: Provides a more fine-grained classification scheme for hardware components such as the target application area for sensors etc. 
	\item Level 3: Is the most detailed specialization of hardware types; sensors, for instance, are further classified according to their measurement principles. 
\end{itemize}

\ifx
\begin{itemize}
\item \textbf{Actuator}: This class represents devices that are operated by a power source and convert energy into motion for interacting with the surrounding environment. While actuator is often merely the term used for a single motor, we use it to describe all possible acting parts of a robot. This class defines subclasses such as \textsf{Robotarm}, \textsf{Mobile Platform}, or \textsf{Conveyor}.

\item \textbf{Controller}: 
This class is used for classifying controllers according to their hardware aspects.
As controllers can be implemented in software only, this class specifically focuses on devices that incorporate some form of integrated processor with controller software implemented running on-board.  
\item \textbf{IO Device}: This class represents digital signal converters. Usually, this means switching a signal in a discrete circuit by a computer or reading out their states. 
\item \textbf{Sensor}: This class represents hardware components that allow a robot to gather information about its state and the environment in which it operates. As depicted in Figure~\ref{fig:hwtypes}, we differentiate between active and passive sensors as well as internal and external sensors. 
Further classification is performed according to the type of output format such as \textsf{Pose}, \textsf{Position}, \textsf{Velocity}, \textsf{Acceleration} etc. Such classifications are calculated by a reasoner based on the \textsf{hasOutputFormat}-property that relates a sensor class to its output format and dimension via subsumption axioms (see~\cite{zander:2015}). 
\item \textbf{HMI Device}\footnote{Abbreviation for Human Machine Interaction Device.}: This class represents those devices that serve as interface between the user and the machine. This hardware type includes types of peripheral devices such as \textsf{Monitor} or \textsf{Joystick}. 
\item \textbf{Object}: This class is used to classify environmental entities that are relevant for a given task or for goals a robot is intended to fulfill. We differentiate between movable and fixed objects.
\item \textbf{Tool}: This class allows for the classification of components that can be used as an end effector of a robot such as welding torches or grippers. 
\end{itemize}
\fi

The hardware ontology also contains axioms that allow for the computation of default capabilities a hardware device of a certain type exhibits.
For instance, the capability of a 2D safety laser scanner can be formally expressed in the following subsumption axiom (cf.~\cite{zander:2015}):
\begin{multline} 
	\label{eq:subsume} 
	\mathsf{SafetyLaserScanner} \sqsubseteq \\ 
	\exists \mathsf{hasCapability.SafeMonitoringOf2DFields}
\end{multline}
The formal interpretation of Axiom~\eqref{eq:subsume} is that if an individual hardware device is classified as \textsf{SafetyLaserScanner} it is \emph{necessary} for it to participate in a \textsf{hasCapability}-relationship to at least one member of the \textsf{SafeMonitoringOf2DFields} class.
This enables a reasoner to infer the capabilities needed, e.g., for building a Pick\&Place application, or to retrieve a list of Apps that satisfy a specific capability requirement. 

\subsubsection{Software Ontology}
\label{ssub:software_ontology}
The software ontology provides a large set of terms for describing various types of software components according to their intended usage (e.g., object detection, control, cognition, planning etc). 
By providing a detailed taxonomical classification system for software components, such information can be used for axiomatically formulating placeholder constraints or to relate default capabilities to software types. 
For instance, the general fact that an object detection software component employs an object detection capability is formulated in the following axiom: 
\begin{multline} 
	\label{eq:swcomp} 
	\mathsf{ObjectDetectionType} \sqsubseteq \\ 
	\exists \mathsf{hasCapability.ObjectDetectionCapability}
\end{multline}
Axiom~\eqref{eq:swcomp} state that the concept  \textsf{ObjectDetectionType} is subsumed by the anonymous class of those things, the members of which participate in at least one \textsf{hasCapability}-relationships to members of the \textsf{ObjectDetectionCapability} class.
Whenever a software component is classified as \textsf{ObjectDetectionType}, it is automatically classified as members of the anonymous super class too.
The software ontology defines classifications on several level that correspond to the different granularity levels in terms of functionality a software component exhibits:
\begin{itemize}
	\item Level 0: Defines the root concept for classifying software components, i.e., \textsf{Software Type}.
	\item Level 1: Represents coarse functionality dimensions such as \textsf{Abstract Planning}, \textsf{Cognition}, \textsf{Control}, \textsf{Human Machine Interaction}, \textsf{Perception}, \textsf{Planning}, \textsf{Simulation}, and \textsf{Coordination}.
	\item Level 2: Represents a more fine-grained classification system being immanent to the specific software type of a software components; for instance, the class \textsf{Control} is classified according to a specific control goal (\textsf{Path following}, \textsf{Tracking}, \textsf{Posture Stabilization}). 
	\item Level 3: Provides the finest classification system; for instance for mapping as a special form of perception, it defines further mapping characteristics such as 2D, 3D, or topological mapping. 
\end{itemize}
In addition, the software ontology is used to retrieve concrete software component instances based on their software type.
\ifx
In the following, we introduce the software ontology's top concepts:

\begin{itemize}
\item \textbf{AbstractPlanning}: Represents the type of software components that enable high-level robot task planning such as \textsf{ActionPlanning} or \textsf{ResourcePlanning}.
\item \textbf{Cognition}: 
Refers to components that exhibit cognitive capacities and allow for adapting a robot's behavior according to situational or environmental changes. 
\item\textbf{Control}: Represents the type of components that regulate the robot movement wrt.\ a specified trajectory from the path planner and sensor states. This class is further classified according to control goals (e.g.~\textsf{PathFollowing} for finding a smooth trajectory where the distance between the mobile robot and the path to follow is minimized).
\item\textbf{Perception}: Represents the type of software components responsible to process and interpret sensor data in order to gain information about properties and elements of the environment. It entails further subclasses such as \textsf{Localization}, \textsf{Mapping}, \textsf{ObjectDetection}, or \textsf{PoseDetection}.
\item\textbf{HMI}: Classifies software components that offer human-machine interaction interfaces to robot systems.
\item\textbf{Simulation}: Used to represent simulation-based software components which can be further classified into mock-up or real simulation. 
\item\textbf{Planning}: Represents those components that generates a trajectory to reach a specified goal. Typically, the output of a planning component is a sequence of intermediate points that form a smooth and continuous trajectory.
\item\textbf{Coordinator}: Represents components that determine the execution sequence of other software components.
\item\textbf{Error Recovery and Exception Handling Type}: Classifies those components the task of which is to compare and detect discrepancies between predictions and observations or perform diagnosis and recovery actions.
\end{itemize}
\fi

\subsubsection{Capabilities Ontology}
\label{ssub:capabilities_ontology}
The capabilities ontology defines a classification system for functionalities HA-Components (or rather their supported hardware devices) and SW-Components are able to perform (e.g., obstacle detection, movement of objects, image processing etc). We understand capabilities as a \emph{metaphysical temporal concept} (cf.~\cite{lenat:1995}) as their existence is determined by the entities they are \emph{qualities of} (cf.~\cite{borgo:2009}). This is the reason why we represent capability information exclusively in the terminological part of ontologies (cf.~\cite{hitzler:2010}), i.e., as TBox axioms rather than as assertional knowledge in the form of ABox axioms\footnote{Representing capabilities as individuals allows for having multiple instances of the same capability which would be factitious and unnatural.} to make full use of the formal semantics of the underlying ontology language (see~\cite{baader:2003,kroetzsch:2014a,rudolph:2011}). 

Capability information is encoded in form of \emph{role restriction axioms} (see~\cite{zander:2015}) where capabilities are linked to a hardware or software type via a \textsf{hasCapability}-role in combination with existential and/or universal quantifiers and the specific capability class as filler. This forms an anonymous superclass that subsumes the concrete type concepts. 

For instance, with Axiom~\eqref{eq:swcomp} and the following capability classification
\begin{multline}
	\mathsf{ObjectDetectionCapability} \sqsubseteq \mathsf{PerceptionCapability}
\end{multline}
the reasoner can deduce that all object detection components also have the default capability \textsf{PerceptionCapability}:
\begin{multline}
	\mathsf{ObjectDetectionType} \sqsubseteq \\
	\exists \mathsf{hasCapability.PerceptionCapability}
\end{multline}
For the materialization of capability information to make them retrievable, capabilities are represented as \emph{DL nominals} (see e.g.~\cite{kroetzsch:2014a,baader:2003,rudolph:2011})  to use them both in ABox and TBox axioms and ensures their \emph{singularity} likewise. 

\section{First Results}

\subsection{Assisted Software Discovery}
\label{sub:using_software}
In this scenario, a system integrator wants to add localization capability to their AGV (automated guided vehicle) platform and has already used the SSMT to find software components of the class \textsf{Localization}.  
The chosen localization software component requires a HA-Component (ROS Wrapper) of the type \textsf{LaserScanner}.
The ReApp Workbench presents all available HA-Components for laser scanners as valid options. Since the integrator already has specific hardware in mind, they further narrow down the results by specifying the manufacturer and model name of their hardware as a \textsf{supported-device} restriction on the HA-Component. In response, the ReApp Workbench reduces the selection to the HA-Components that support this particular device. In addition, it takes the specific requirements for the application into account to limit the selection: For example, the selected localization component might require a certain update frequency from the laser scanner and also expect to use the measured reflectance values (which currently not all  ROS drivers provide). Only the HA-Components that provide the required features are shown as suitable candidates.
The system integrator finally makes their choice---according to external criteria like ratings or price---and can be sure that it can be used with the localization software component.

With the ontology-based approach used in ReApp, such a requirement can be expressed in form of a subsumption axiom, i.e., it is reformulated as an anonymous complex class expression in form of a DL query that subsumes those classes, the members of which satisfy the given class restriction. 
The membership of eligible individuals is computed automatically by the DL reasoner. 
For the given example, the requirement is formulated as the following DL query\footnote{We have omitted namespaces for readability issues; all terms used in the DL Query are defined in the ReApp ontologies.}:
\begin{multline}
	\mathsf{HAComponent} \sqcap \exists \mathsf{supportsDevice.(LaserScanner} \; \sqcap \\ 
	\exists \mathsf{hasAttribute.(UpdateFrequencyInHz} \; \sqcap \\
	\exists \mathsf{hasIntValue.}(>=,30)) \; \sqcap \\ \exists \mathsf{hasAttribute.MeasuredReflectance}) 
\end{multline}
System users, of course, do not have the specific technical knowledge to formulate such statements. Therefore, we have developed a Web-based user interface for the ReApp Store that guides users in formulating requirements for software components. More information on this is given in~\cite{bastinos:2014}.

\subsection{Generating Semantic Annotations}

\label{sub:writing_software}
In this scenario, a software developer is tasked to develop a drop-in replacement of the localization software component from the previous scenario since this component did not perform well in the AGV's particular environment.

When the developer selects the classes \textsf{Localization} and \textsf{2D}, the ReApp Workbench automatically adds the associated capabilities and attributes to the semantic model of the new software component. 
This includes the localization capability for the two dimensional case as well as corresponding ROS interfaces---in this case, a topic with message type \textsf{geometry\_msgs/Pose2D}.
The developer then adds additional capabilities they want to implement. They also define that the new software component requires a HA-Component of class \textsf{LaserScanner} with a certain resolution and frequency---however, without restricting the component to require a specific model of laser scanner.
The new localization component can now replace the previously used one as it realizes the same functionality (localization) and provides the same interface to other components. Thus making both localization components interchangeable. 
The semantic annotations are also used during implementation and testing: The engineering tool uses the information to generate a code skeleton into which application-specific code is inserted. Additionally, the new component is automatically tested to whether its interfaces and capabilities are compliant with domain semantics before it is published.

\subsection{Modularizing Complex Operations Sequences}
\label{sub:bmw_use_case}
For this pilot demonstrator, an assembly task was chosen which is highly repetitive and must be executed with precision, making it an ideal task for a robot. A reference solution for this task, implemented conventionally, is already in place at BMW, a consortium member.
The ReApp based implementation of this traditional automation scenario will showcase how ReApp can make software development more dynamic, flexible, and modular.

\begin{figure}[ht]
	\centering
	\includegraphics[width=\linewidth]{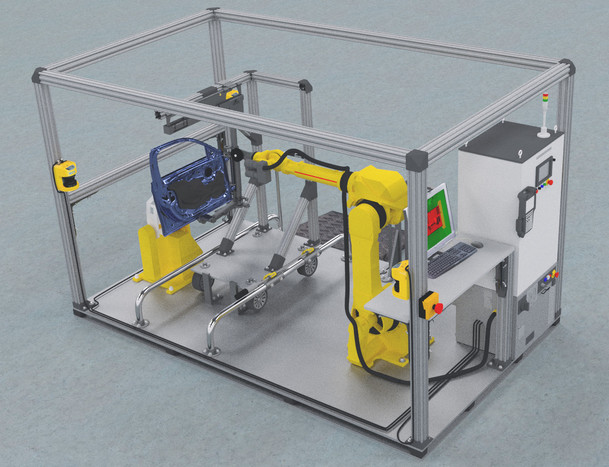} 
	\caption{The demonstrator covers a traditional automation scenario with multiple interchangeable skills to provide flexible solutions}
	\label{fig:BMW_use_case}
\end{figure}

The assembly task consists of pressing sound insulation matting against a variety of car doors. The sound insulation matting, which is specific for each type of door, is prepared with adhesive around its contours and loosely fitted into the door frame by a human worker. 
As the adhesive requires a consistent pressure to harden reliably, a robotic arm is used to follow the contours of the insulation matting while exerting a constant force with its pressing tool.
While the traditional implementation satisfies all process requirements, some parts such as the door localization take up a considerable amount of time and the trajectory for each door has to be taught manually in a time consuming process by robotic experts.
Most importantly, exchanging parts of the software solution (such as the door localization) or adding a new trajectory requires the system integrator to rewrite at least parts of the code. Due to the robot manufacturer-specific language, the program is also not reusable for other robots.
The proposed ReApp solution to this problem subdivides the task into multiple reusable and interchangeable skills:

\begin{itemize}
\item Door/Object Detection Skill
\item Door/Object Localization Skill
\item Trajectory Execution Skill
\item Safety Control Skill
\item Trajectory Teach-In/Import Skill 
\end{itemize}

Each part is implemented with an equivalent to the current solution, but can easily be replaced with an alternative approach.
For example, the demonstrator setup also allows door localization to be implemented using an integrated vision system provided by the robot manufacturer; another alternative is an off-the-shelf (stereo) camera or depth camera in combination with software for object detection.
By following the proposed modelling methods, all these variants of the ``Door/Object Localization Skill'' will present the same interfaces and capabilities required by the other high level components, making them interchangeable within the system without having to re-engineer the whole solution.

\ifx

Besides showing that these different solutions can be easily interchanged, our aim is also to show that traditional automation solutions, such as integrated sensing, can co-exist with approaches more readily associated with novel and academic applications such as depth image evaluation. 
Which one is suitable for the task depends on the resources of the project---which will have an impact on the hardware that is available. Since dependencies on hardware can be modelled in ReApp, they can automatically be evaluated, helping the creator of the application to select a suitable solution without extended background knowledge in robotics or detailed information about the individual skills.

\fi
By choosing an industrial application with an existing reference implementation, this use case offers the chance to compare the ReApp solution with the traditional approach---with respect to runtime behaviour, but especially regarding the expected benefit in terms of engineering time for setup and modification.  

\section{DISCUSSION AND CONCLUSION}
In this work, we introduced the ReApp approach to reusable software components.
ReApp provides a toolset 
for model-driven engineering which is seamlessly integrated with ontologies that ease the process by providing structural and domain knowledge.
We have presented the underlying information model and the ontologies it is based upon and detailed the workflow for creating and retrieving software components.
To demonstrate the viability of our approach, the workflow for different user roles and the expected benefit for larger applications has been presented.

The ReApp tool chain and its ontology-based information model enable the model-driven engineering of components while aiding the user with the knowledge encoded within the ReApp domain ontologies. They therefore greatly enhance the chance of software reuse, lifting the burden of having to be proficient with every piece of software.
Such enhanced usability and especially reusability of software enables faster developments of custom robotic applications and, most importantly, keeps them maintainable to allow flexible response to market demands.

Future work focuses on further integrating constituting tools, such as the test framework in the ReApp Store for automatic functionality performance evaluation of uploaded Apps, and improving the ReApp ontologies towards the integration of externally hosted data to link component semantics with well-known ontologies in the Web of Data to foster interoperability and to implement advanced verification and validation algorithms.

\addtolength{\textheight}{-12cm}   

\section*{ACKNOWLEDGMENT}
This work has been partially funded by the German Federal Ministry for Economic Affairs and Energy through the project ReApp (no. 01MA13001).


\bibliography{refs}{}
\bibliographystyle{ieeetr}

\end{document}